\documentclass[conference]{IEEEtran}
\IEEEoverridecommandlockouts
\usepackage{cite}
\usepackage{amsmath,amssymb,amsfonts,mathrsfs}
\usepackage{algorithmic}
\usepackage[linesnumbered,ruled]{algorithm2e}
\usepackage{graphicx}
\usepackage{textcomp}
\usepackage{multirow}
\usepackage{xcolor}
\usepackage{bm} 
\usepackage{enumitem}
\usepackage{dsfont}
\usepackage{subcaption}
\usepackage{url}
\usepackage{booktabs}
\usepackage{xspace}
\def\BibTeX{{\rm B\kern-.05em{\sc i\kern-.025em b}\kern-.08em
    T\kern-.1667em\lower.7ex\hbox{E}\kern-.125emX}}
\begin{document}
\newcommand{\ie}{i.e.,\xspace}
\newcommand{\eg}{e.g.,\xspace}

\title{You Don't Need All Attentions: Distributed Dynamic Fine-Tuning for Foundation Models\\
}

\author{\IEEEauthorblockN{Shiwei Ding\IEEEauthorrefmark{2}, Lan Zhang\IEEEauthorrefmark{3}, Zhenlin Wang\IEEEauthorrefmark{2}, Giuseppe Ateniese\IEEEauthorrefmark{4}, Xiaoyong Yuan\IEEEauthorrefmark{3}}
\IEEEauthorblockA{\IEEEauthorrefmark{2}Michigan Technological University}
\IEEEauthorblockA{\IEEEauthorrefmark{3}Clemson University}
\IEEEauthorblockA{\IEEEauthorrefmark{4}George Mason University}
\IEEEauthorblockA{shiweid@mtu.edu, lan7@clemson.edu, zlwang@mtu.edu, ateniese@gmu.edu, xiaoyon@clemson.edu}
}
\maketitle
\begin{abstract}
Fine-tuning plays a crucial role in adapting models to downstream tasks with minimal training efforts. 
% With minimal training effort, time, and data, fine-tuning allows pre-trained models, especially large foundation models, to be efficiently and effectively tailored for specific applications. 
% With minimal training effort, time, and data, fine-tuning allows pre-trained networks—especially large foundation models—to be efficiently and effectively tailored for specific applications. 
However, the rapidly increasing size of foundation models poses a daunting challenge for accommodating foundation model fine-tuning in most commercial devices, which often have limited memory bandwidth. Techniques like model sharding and tensor parallelism address this issue by distributing computation across multiple devices to meet memory requirements.
% split the neural network into small subnets while deploying subnets on distributed devices to meet memory constraints. 
Nevertheless, these methods do not fully leverage their foundation nature in facilitating the fine-tuning process, resulting in high computational costs and imbalanced workloads. 
We introduce a novel Distributed Dynamic Fine-Tuning (D2FT) framework that strategically orchestrates operations across attention modules based on our observation that not all attention modules are necessary for forward and backward propagation in fine-tuning foundation models. 
% We creatively observe that not all attention modules are needed for performing forward and backward propagation operations in fine-tuning foundation models.
% Therefore, we develop a Distributed Dynamic Fine-Tuning (D2FT) framework that dynamically orchestrates the operations on attention modules. 
Through three innovative selection strategies, D2FT significantly reduces the computational workload required for fine-tuning foundation models. Furthermore, D2FT addresses workload imbalances in distributed computing environments by optimizing these selection strategies via multiple knapsack optimization.
% Via three innovative selection strategies, this framework significantly reduces the computational workload for fine-tuning foundation models. 
% Additionally, D2FT addresses workload imbalance in distributed computing by optimizing the selection strategies via multiple knapsack optimization.
Our experimental results demonstrate that the proposed D2FT framework reduces the training computational costs by $40\%$ and training communication costs by $50\%$ with only $1\%$ to $2\%$ accuracy drops on the CIFAR-10, CIFAR-100, and Stanford Cars datasets.
Moreover, the results show that D2FT can be effectively extended to recent LoRA, a state-of-the-art parameter-efficient fine-tuning technique. By reducing $40\%$ computational cost or $50\%$ communication cost, D2FT LoRA top-1 accuracy only drops $4\%$ to $6\%$ on Stanford Cars dataset. 
\end{abstract}
% \vspace{-1em}
\section{Introduction}
\label{sec:intro}
Foundation models are extensively trained on vast datasets that cover a wide array of topics and knowledge domains, thereby facilitating a deep understanding of real-world concepts~\cite{devlin2018bert, liu2019roberta, raffel2020exploring, yang2019xlnet, sanh2019distilbert}. Leveraging this foundational understanding, these models can be efficiently adapted to practical machine-learning tasks by fine-tuning their parameters from the pre-trained foundation models. The fine-tuning uses fewer training epochs and fewer data inputs to achieve a low-cost adaptation.
% However, the enormous sizes of foundation models (e.g., GPT-4~\cite{bubeck2023sparks} with 1.8 trillion parameters, Vision Transformer~\cite{yuan2021tokens} with 86 million parameters) present a daunting challenge for fine-tuning. 
% Since these models are already capable in a general sense, they can be fine-tuned with additional, smaller datasets to excel at specific tasks, i.e., downstream tasks. 
However, fine-tuning a foundation model still demands significant memory resources to store model parameters (e.g., GPT-4~\cite{bubeck2023sparks} with 1.8 trillion parameters, Vision Transformer~\cite{yuan2021tokens} with 86 million parameters), intermediate activations, and gradient updates, within GPU memory. 
% This creates a significant challenge for most commercial devices, such as consumer-grade GPUs and cloud ML services, as their limited memory capacity struggles to accommodate the increased size of large foundation models. 
This creates a tension between the growing size of neural networks and the memory constraints of commercial devices such as consumer-grade GPUs and cloud ML services, limiting the scalability potential of foundation models~\cite{li2020train}. 
% as recent empirical findings indicate that increasing the number of parameters and layers generally improves model performance~\cite{li2020train}. 
%Nevertheless, recent empirical findings have also shown that an increase in the number of parameters and layers typically enhances model performance~\cite{li2020train}. 
%It is anticipated that the tension between the increasing size of neural networks and memory constraints will continue to be a prevalent trend, potentially exacerbated in the fine-tuning of foundation models, thereby hindering the full utilization of these models' capabilities.

\begin{figure}[!tb]
    \centering
    \includegraphics[width=0.99\linewidth]{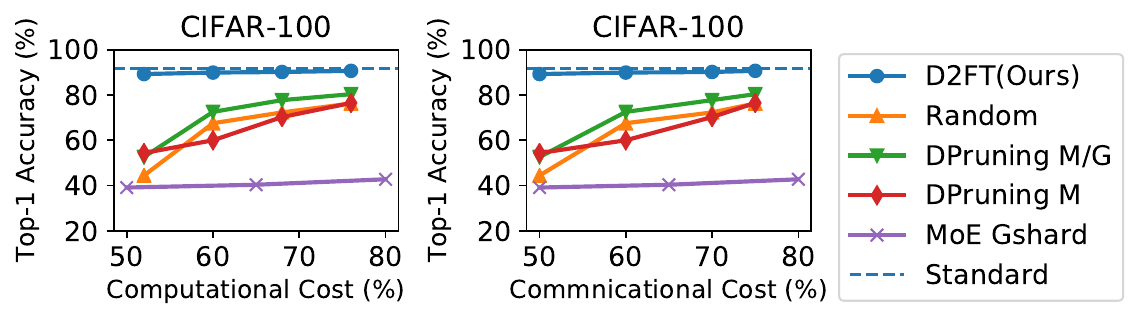}
    % \caption{The full fine-tuning top-1 accuracy comparison under the same or similar computation and communication costs on Cifar-100 datasets.
    %From the lines shown in this figure, our method outperforms the baseline methods on Cifar-100 and Cifar-10 when the communication costs are close.
    % }
    % \label{fig: Experiment results Cifar-100}
% \end{figure}
% \begin{figure}[!tb]
    % \centering
    \includegraphics[width=0.99\linewidth]{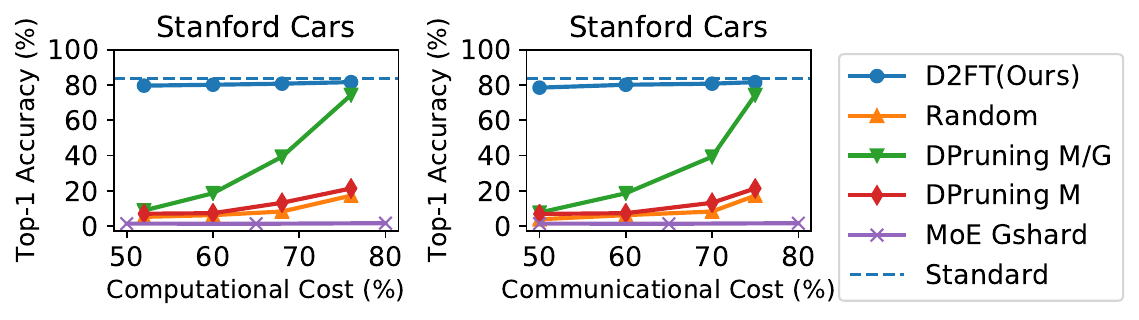}
    \caption{(Full parameter) Fine-tuning performance comparison. The proposed D2FT framework outperforms existing efficient distributed learning frameworks under similar computation and communication costs on the CIFAR-100 and Stanford Cars datasets. The ``Standard'' means the model standard full fine-tuning.
    % The full fine-tuning top-1 accuracy comparison under the same or similar computation and communication costs on Stanford Cars datasets.
    %From the lines shown in this figure, our method outperforms the baseline methods on Cifar-100 and Cifar-10 when the communication costs are close.
    }
    \label{fig: Experiment results Car}
    %\vspace{-1.2em}
\end{figure}

To address these challenges, distributed learning paradigms like pipeline parallelism~\cite{belilovsky2020decoupled, yuan2022distributed, yuan2022decentralized} and tensor parallelism~\cite{shoeybi2019megatron, xu2021gspmd, xu2023efficient, bian2021maximizing} have become widely adopted as effective solutions.
%\by{update model sharding to pipeline parallelism, see https://huggingface.co/transformers/v4.10.1/parallelism.html}
%\by{add tensor parallelism.}
In pipeline parallelism, the neural network is divided into layer-wise segments, with each segment deployed on a different device, enabling efficient use of distributed resources. Tensor parallelism, by contrast, partitions individual tensors (weight matrices) into chunks, distributing each chunk across separate devices. These methods help balance workloads by distributing parameters across devices, reducing both memory and computational demands on any single device.
% train a neural network over distributed devices. 
% The layers or blocks of neural networks are partitioned into subnet works, thereby . 
% In this paper, we focus on Transformers.
For instance, the vision transformer~\cite{vaswani2017attention,dosovitskiy2020image}, a widely used foundation model architecture, can be split by blocks or by attention heads (in each block) and then distributed across devices for accommodation.

While distributed training methods offer the advantage of memory saving, they are mainly investigated in the traditional training paradigm, where every model parameter is involved in all training operations.
% receive all training data. 
This line of research overlooks the unique potential of the foundation models, whose parameters have already encoded extensive (and somewhat redundant) prior knowledge that can facilitate more efficient fine-tuning. 
% It has been observed that it is effective to adapt to downstream tasks without fine-tuning all the parameters in the foundation model~\cite{touvron2022three}. 
% As one key benefit of large-scale neural networks, to fine-tune a general-purpose foundation model on a downstream task, it is not necessary to update all the parameters~\cite{touvron2022three}. 
% This intuition inspired our initial design to reduce the workload for certain modules of the model.
Inspired by this insight, we aim to design an innovative framework to selectively reduce the fine-tuning workload on foundation models.
% We hypothesize that given a downstream training sample, no
We observe that not all the attention modules in both forward pass and backward pass significantly contribute to fine-tuning a given training sample. Therefore, we can skip the operations on the specific subnet during fine-tuning, which largely saves computational and communicational costs while preserving the fine-tuning accuracy.
% First, to adapt to a specific downstream task, not all the parameters are required to be updated. Hence, the backward operations could be skipped on a specific subnet. 
% First, in the backward pass, it has been shown that not all the parameters and their gradients are required to be calculated and updated. Hence, we can skip the backward operations on specific subnet.
% First, given a downstream training sample, in the backpropagation, gradients of a part of subnet do not need to be computed. 
% , which largely saves the computational, communication, and energy costs.
% Second, in the forward pass, our empirical results show that given a training sample, a specific subnet can be approximated by a linear function without affecting its performance. 
% Then, the forward operations on these subnets (and their devices) can also be skipped, with minimal cost of running the linear function on the other devices.
Given a training sample, we consider three operation options for each subnet: i) performing both forward and backward operations, ii) performing the forward pass only, and iii) skipping operations entirely, detailed in Section~\ref{sec:operation selection}.
However, applying these operations may lead to two issues. First, fine-tuning performance drops as the foundation model does not fully learn the new knowledge due to the missing learning operations. Second and more importantly,  workloads are significantly imbalanced among devices as some subnets play more critical roles than others, resulting in heavier workloads on their corresponding devices, and significantly slowing down the entire system.
% since a few subnets are more important than others, and more intensive workloads will be allocated to their corresponding devices. 
% The devices with important subnets will then suffer from more intensive loads and become bottlenecks, significantly slowing down the entire system.
To tackle these challenges, we propose Distributed Dynamic Fine-Tuning (D2FT), which dynamically allocates learning operations across distributed devices for fine-tuning a foundation model. 
Through three innovative selection strategies, D2FT maintains the fine-tuning accuracy while significantly reducing the computational workload required for fine-tuning foundation models. Furthermore,
D2FT addresses workload imbalance in distributed settings
by optimizing selection strategies through multiple knapsack
optimization, a classic combinatorial problem that maximizes
value under capacity constraints. 
% Specifically, we first partition a neural network into multiple attention-based subnets and assign them to individual devices, ensuring each subnet fits within the device's memory capacity.
% Accordingly, we formulate the scheduling as a multi-knapsack problem. 
% D2FT is designed to optimize learning performance without overloading the workload on any device by dynamically selecting appropriate learning options. 
Additionally, we further extend D2FT to low-rank adaptation (LoRA)~\cite{hu2021lora}, a state-of-the-art parameter-efficient fine-tuning technique~\cite{houlsby2019parameter}. The experimental results demonstrate the effectiveness of D2FT in both full-parameter and parameter-efficient fine-tuning.
Therefore, our major contributions are four-fold:

\begin{itemize}[leftmargin=*]
\item We develop a Distributed Dynamic Fine-Tuning (D2FT) framework that strategically orchestrates attention operations across devices to address high computational and communicational costs in distributed foundation model fine-tuning. 
% that is memory and computation-friendly for recourse constraint devices.

\item We formulate the orchestration problem by introducing three innovative selection strategies and addressing it through a hierarchical decoupling of the multi-knapsack problem.
% to address workload imbalances in distributed computing environments.

\item We extend D2FT to LoRA fine-tuning by separately allocating low-rank matrices with frozen attention heads and feed-forward networks across different devices. %\by{add a bit of details of our methods on LoRA}.
% select important samples (micro-batch) to process for each subnet, reducing computation and communication costs over devices.

% \item We formulate a multi-knapsack problem for the load-balancing challenge and derive the solution.  

\item The experimental results show that D2FT outperforms baseline methods, reducing the computational cost of training (fine-tuning) by $40\%$ and the cost of training communication by $50\%$ with a decrease in precision $1\%$ and $2\%$ in the CIFAR-10 and CIFAR-100 datasets, respectively. In addition, the experimental results on LoRA show that D2FT can reduce LoRA computational cost by $40\%$ and communicational cost $50\%$, respectively, with $4\%$ and $6\%$ accuracy drop, respectively. 

%communication costs of tensors transformation with merely $0.9\%$ and $5.2\%$ top-1 accuracy drop on the Cifar-10 and Cifar-100 datasets.  \by{double check these numbers, write in this way: reduce the xx cost by xx\% ... see abstract}
\end{itemize}
% \vspace{-0.3em}
%\vspace{-0.7em}
\section{D2FT Framework}
\label{sec:method}
%\vspace{-0.5em}
%This paper develops a scheduling strategy to optimize learning operations across devices, reducing costs without sacrificing accuracy or workload balance.

%\subsection{Problem formulation}
%\vspace{-0.3em}
\subsection{Problem Formulation}
\label{sec:Formulation}
D2FT aims to derive optimal orchestration across multiple devices to meet memory constraints while achieving computational and communication efficiency in fine-tuning foundation models.
Following the existing work, we first formulate the orchestration problem.
Suppose a large-scale foundation model $\mathcal{F}$ is fine-tuned on a downstream dataset $\mathcal{D}=\{\bm{x}_i\}$ with $N$ samples.
%To fit the large scale of a foundation model to each device, 
To distribute computation over $K$ devices, the foundation model $\mathcal{F}$ is partitioned by $K$ subnets: $\mathcal{F}_{1}, \cdots, \mathcal{F}_{k}, \cdots, \mathcal{F}_{K}$ and deployed across $K$ devices, where each subnet can be fit into the device\footnote{For simplicity, we assume that the number of subnets is set as the number of devices, which can be adjusted according to model sizes and device capacities.}. %\by{add a simple solution here: For example, ...}. }.

% Our key idea is to schedule all inputs, assign them to every subnet module according to prediction scores, and ensure that each subnet module receives the same number of inputs in the full fine-tuning process. Therefore, we formulate the scheduling task as the following optimization problem. 
% This orchestration problem can be formulated as follows. 
Unfortunately, existing work mainly distributes computation across devices while not fully leveraging the inherent capability of foundation models - model parameters have already included extensive and redundant prior knowledge. Thus, a large number of operations in fine-tuning are not fully essential. 
To fully unleash the power of foundation models, we innovatively introduce an innovative and diverse set of operation strategies to avoid unnecessary computation and communication. Specifically, for a sample $\bm{x}_i$, each device $k$ can select an operation $\mathscr{p}$ from the operation set $\mathcal{P}$,
% (\eg, forward only, forward and backward, shortcut, 
detailed in the Section~\ref{sec:operation selection}. 
Our goal is to optimize the selection strategies to achieve optimal fine-tuning performance within the given workload constraints in devices. Since it is non-trivial to measure the fine-tuning performance, we use a proxy measurement, data contribution, detailed in Section~\ref{sec: scores full}.
We denote $\mathcal{A}^{p}(\mathcal{F}_{k}), p \in \mathcal{P}$ as the contribution of selecting operation $\mathscr{p}$ on the $k$-th subnet when training on the $i$-th data sample.
% Given the $k$-th subnet and the $i$-th data sample, we denote $\mathcal{A}^{\mathscr{p}}(\mathcal{F}_{k}), \mathscr{p} \in \mathcal{P}$ as its contribution of selecting operation $\mathscr{p}$.  
%\by{the notation of p has been updated. please update the following text.} 
Additionally, given an operation $p$, the computational cost is denoted as $\mathcal{C}^{p}(\mathcal{F}_{k})$. 
% Our goal is to maximize the contributions to the training accuracy (\ie, achieving optimal learning) while maintaining the workload balance.
Hence, the orchestration problem can be formulated as:
% \begin{equation}
% \max_{p_k} \sum_k^K  \left( \sum_i^N\mathcal{A}^{p_k}(\mathcal{F}_{k}) - \lambda \| \sum_i^N \mathcal{C}^{p}(\mathcal{F}_{k}) - \bm{C}_k\|\right) 
% \end{equation}
% where $\bm{C}_k$ denotes workload expectation of the $k$-th device. 
% $\bm{C}_k$ is calculated based on the computational capacity of each device. 
% The first term aims to maximize the contribution scores, \ie, learning performance, while the second term aims to achieve balanced workloads across devices by reducing workload variances. 

% \by{where is p?}
\begin{align}
\label{eq:optimization}
    \max_{p} \quad & \sum_k^K\sum_i^N \mathcal{A}^{p}(\mathcal{F}_{k}) \\
    \mathrm{s.t.} \quad & \sum_i^N\mathcal{C}^{p}(\mathcal{F}_{k}) \le \bm{C}_k \quad (\forall k), \nonumber
\end{align}
where $\bm{C}_k$ denotes workload expectation of the $k$-th device. $\bm{C}_k$ is calculated based on the computational capacity of each device. 
The objective function aims to maximize the contribution scores, i.e., learning performance, while the constraints ensure balanced workloads, i.e., no straggler devices. The imbalance of workloads can be measured as minimizing its Lagrange term $\| \sum_i^N \mathcal{C}^{p}(\mathcal{F}_{k}) - \bm{C}_k\|$.

% $\bm{C}_k$ aims to achieve two goals. First, $\bm{C}_k$ is designed to reduce the computational cost of each device. By adopting a small $\bm{C}_k$, more lightweight operations will be selected in the learning process. 
% \by{logic flow is not good here - why using constraint can achieve balanced workloads? still not quite convincing here.} 
% Second, when all device expectations $\bm{C}_k$ are equal, we could ensure the workload balancing across devices. 
% Second, since no intensive workload beyond the workload constraint $C_k$ will be assigned to individual devices, $\bm{C}_k$ ensures the workload balancing across devices. 
% If all devices possess identical computational capacities, the same value will be assigned to $C_k$. 

To derive the optimal solutions to the orchestration problem, in the following subsection, we begin by introducing the general partitioning strategy for foundation models (Section~\ref{sec: partition full}), followed by a description of the proposed operation selections (Section~\ref{sec:operation selection}). Next, we detail the measurement of contribution scores (Section~\ref{sec: scores full}) and conclude with a summary of the overall optimization problem (Section~\ref{sec: optimization problem}).
% We aim to address the following questions in optimizing Eq.~\ref{eq:optimization}: \by{add section number for each question}
% How should the foundation model be partitioned? \by{is this a novel design? if not, just move Section 3.1 to the experimental settings.}
% How should the operation set $\mathcal{P}$ be selected?
% How can we calculate each operation's contribution to training accuracy?
% How can we best optimize Eq.~\ref{eq:optimization}?
%\vspace{-0.3em}
\subsubsection{Partitioning strategy}
%\vspace{-0.3em}
% For a large-scale neural network, even one layer or block is large enough for most devices. It is not enough to solve a severe lack of resources case if we only consider block-wise (separate the model by layers or blocks) or width-wise(separate the model by tensors). Therefore, we split the model in both depth-wise and width-wise.

% We focus on separating the transformer base network, which this structure is popularly used in large-scale networks. For example, the transformer-based network contains $12$ blocks, and each block contains $6$ multi-attention heads with a feed-forward network(FFN). We can split the network into $K = L \times H$ parts with each partitioned subnet $\mathcal{F}_{k}$ contains $12\mid L$ blocks with $6\mid H$ heads and $\frac{1}{H}$ part of FFN, where the $L$ and $H$ is divisible by $12$ and $6$ respectively. To have a better illustration, we use $k$ to index the subnet, where $k = \{l,h\}, l\in L, h \in H$. 
%\by{move this to operation selection}
%For each subnet $\mathcal{F}_{k}$, we set a residual route (bypass route) to send the previous results/gradients to the latter subnets if this subnet is not chosen. This pass will not affected by the operation selection of each subnet.
\label{sec: partition strategy}
In this work, we focus on transformer-based architectures~\cite{han2021transformer} due to their prevalence in foundation models. Transformer architectures generally consist of $L$ blocks, each containing $H$ attention heads and a feed-forward neural network. Building on recent advances in model sharding and tensor parallelism~\cite{shoeybi2019megatron, yuan2022decentralized}, we apply depth-wise and width-wise partitioning to the transformer to derive subnets.
The minimal unit of a subnet module is one attention head with a small feed-forward network (FFN) in a block. The proposed framework can be easily extended to devices consisting of multiple attention heads and FFNs.
%The dimension of the network is reduced by $H$ times compared with the original neural network. 
We denote each subnet module as $\mathcal{F}_k$, where $k=(l,h), 1\leq l\leq L, 1\leq h\leq H$, where $l$ and $h$ denotes the index of attention heads and index of blocks in the foundation model.
\label{sec: partition full}
%\vspace{-0.2em}
\subsubsection{Operation selection}
%\vspace{-0.3em}
\label{sec:operation selection}
%\by{explain why these three types of operations?}
To solve the orchestration problem, we introduce a set of operations $\mathcal{P}$ that reshapes the learning paradigms for fine-tuning foundation models, detailed as follows: 
% , making it crucial to select operations that maintain learning performance while improving workload balance. 
% In this paper, we identify three operations, 
\begin{itemize}[leftmargin=1em]
    \item Full operation (${p_f}$). This operation aligns with the standard training paradigm in neural networks, where both forward and backward passes are conducted during training.
    %\by{explain in detail of this operation, we have many things to discuss, how to skip in forward pass, how to skip in backward pass, the current description is very superficial} 
    % The subnet with Forward-Backward operation shall propagate forward and backward and update itself. 
    %The subnet with Forward-Skip operation shall choose the residual route to bypass the subnet in forward propagation, and thus, there will be no updates with this part of the network.
    \item Shortcut operation (${p_s}$). Motivated by recent work in sparse training~\cite{lin2020dynamic, kohama2023single}, we introduce the Forward-Skip (FS) operation, which bypasses all operations on a specific subnet, including both the forward and backward propagation. To minimize the impact on the rest of the operations, a shortcut is introduced to propagate activations and gradients.
    % As a result, the subnet module does not update. 
    % When  FS is activated, previous results and backward gradients are routed through a residual path to the next and preceding subnets, respectively. 
    \item Forward-Only operation (${p_o}$). 
    %\by{the name sounds similar to Forward-Skip, let's use Full operation $p_f$, Shortcut operation $p_s$, and Forward-Only operation $p_o$? Please update the remaining text.} 
    Bypassing forward propagation may significantly affect the model's performance since the activations are not accurately calculated. Therefore, we introduce the Forward-Only operation that remains the forward propagation but does not update the gradients, i.e., skipping backward propagation. %\by{verify this}
\end{itemize}
Note that we establish a residual route for each subnet module to ensure that forward or backward propagation is not disrupted. This route propagates activations or gradients to subsequent subnets if an $p_s$ or $p_o$ operation is selected, ensuring the integrity of the learning process.
\subsubsection{Contribution scores}
%\vspace{-0.4em}
Given the intractable impact of different operations on learning performance, we introduce two innovative proxy measurements: i)  backward contribution score and ii) forward contribution score. These scores respectively quantify a subnet's contribution to backward propagation and forward propagation.
Note that as data samples strongly influence the importance of each subnet, the contribution scores vary across different data samples. 
% i) backward contribution score for determining the Forward-Backward (FB) operation and ii) forward contribution score, for determining the Forward-Pass (FP) operation adopted in a subnet.
% Calculating the contributions of different operations to the model performance is challenging and intractable. 
% We design two contribution scores, forward-score and backward-score, to estimate whether a subnet should use the $FP$ or $FB$ option.\\
% We propose two contribution scores, \ie, backward contribution score for determining the Forward-Backward (FB) operation and forward contribution score, for determining the Forward-Pass (FP) operation adopted in a subnet.
% \by{do we also have other options in the results? Write it in a more general way, and then provide all contribution score options.}
To establish an effective measurement approach, we explore commonly used metrics for assessing sample contribution, including Fisher Information, Gradient Magnitude, Weight Magnitude, and Taylor Expansion-based Weight Importance. 
We explore the effectiveness of these metrics and empirically select Fisher Information and Weight Magnitude as forward and backward contribution scores, which are defined as follows.

% \begin{itemize}
\noindent \textit{Fisher Information}: The contribution score of using operation $p$ sums up all weights' empirical Fisher information in a subnet $\mathcal{F}_k$, following~\cite{ilhan2024resource}: 
\begin{equation}
{\mathcal{A}}_k^p = \sum \|\nabla_{w_k}\|^2,
\end{equation}
where $\nabla_w$ calculates the gradients of subnet $\mathcal{F}_k$ with respect to its weights $w_k$ a given sample $x_i$.

\noindent \textit{Weight Magnitude}: The contribution score sums up all weights' magnitude for a subnet $\mathcal{F}_k$:
\begin{equation}
\mathcal{A}_k^p = \sum\| w_k\|.
\end{equation}
Note that to calculate these scores, we feed all samples for forward and backward propagation without updating weights to calculate all subnets' Fisher information before fine-tuning. We record the magnitude of all pre-trained subnets to calculate each subnet's Weight Magnitude. 
The comparison with other metrics and their definitions can be found in ablation studies (Section~\ref{sec: score candidate}).

\label{sec: scores full}
%\vspace{-0.2em}
\subsubsection{Overall orchestration problem}
%\vspace{-0.3em}
\label{sec: optimization problem}
%Given the selection of operations and their contribution scores, this section summarizes the optimization problem.
%In evaluating neural network models on datasets, the computation cost for each forward propagation step is denoted as $c$, while the cost for backward propagation is represented by $d$. 
%We also impose a mutually exclusive constraint on each data sample $x_i$: each subnet can only choose one possible operation selection. 
%Our scheduling objective is twofold: to achieve workload balance across all subnets by ensuring uniform computation costs among devices (subnets) while maximizing the contribution scores for each subnet. This optimization problem is akin to the multi-knapsack problem~\cite{martello1990knapsack}, where there exists a total of $K$ knapsacks (devices), each with a specific ``weight capacity'' $C_k$. The aim is to maximize the overall ``value'' $V$ (contribution score) while adhering to the constraints imposed by the weight capacities of the knapsacks. Therefore, our optimization multi-knapsack problem can be formulated as follows:
We summarize the orchestration problem based on the aforementioned partitioning strategy, operation selection, and contribution scores as follows:
%\by{check the equation $p_o?$ }
\begin{align}
\label{eq:overall_problem}
\max &  \sum_k^{K}\sum_{i}^{N}\mathds{1}_{\mathscr{p}p_f}(x_i) \mathcal{A}^{p_f}(\mathcal{F}_k) + \mathds{1}_{\mathscr{p}p_o}(x_i) \mathcal{A}^{p_o}(\mathcal{F}_k)\\
s.t. &  \sum_{i}^{N} \mathds{1}_{\mathscr{p}p_f}(x_i)(c^f_k + c^b_k) + \mathds{1}_{\mathscr{p}p_o}(x_i) c^f_k \leq \bm{C}_k,\nonumber\\
& \mathds{1}_{\mathscr{p}p_f}(x_i) + \mathds{1}_{\mathscr{p}p_o}(x_i) \leq 1, \\
& \mathds{1}_{\mathscr{p}p_f}(x_i), \mathds{1}_{\mathscr{p}p_o}(x_i) \in \{0,1\}.\nonumber
\end{align}
$c^f_k$ and $c^b_k$ denote the computational costs in forward and backward propagation for a subnet $k$, respectively. 
%\by{is c and d relevant to the subnet? can we add the subnet info? like $\mathcal{C}_k^f$, $\mathcal{C}_k^b$, extended to communication cost?}
$C_k$ is the computational cost of the $k$-th device. $\mathds{1}_{\mathscr{p}p_f}(x_i)$ and $\mathds{1}_{\mathscr{p}p_o}(x_i)$ %\by{remove parenthesis: $\mathds{1}_{\mathscr{p}{p_f}}^{x_i}$} 
are the indicators of the operation selections for the sample $x_i$. $\mathds{1}_{\mathscr{p}p_f}(x_i) =1$ if the full operation is selected, $\mathds{1}_{\mathscr{p}p_o}(x_i) =1$ if the forward-only operation is selected, otherwise, the shortcut operation $p_s$ is selected. The selection of full operation and the forward-only operation conflicts for each subnet, i.e., $\mathds{1}_{\mathscr{p}p_f}(x_i) +\mathds{1}_{\mathscr{p}p_o}(x_i) \le 1$. 
\subsection{Proposed Heuristic Algorithm}
\label{sec:Heuristic Algorithm}
% \vspace{-0.3em}
% \input{sec/Algorithm_new}
% \input{sec/Algorithm_2new}

This orchestration problem (Eq.~\ref{eq:overall_problem}) resembles a classic multi-knapsack problem~\cite{martello1990knapsack}, in which items (in this case, subnets or devices) are selected to optimize a certain value (contribution scores) without exceeding capacity constraints (computational costs) of multiple ``knapsacks'' or resources. However, it is intractable to solve this problem due to two key reasons. First, the multi-knapsack problem is, unfortunately, NP-hard~\cite{kan1993class}. Second, although the contribution scores $\mathcal{A}^{p_f}(\mathcal{F}_k), \mathcal{A}^{p_o}(\mathcal{F}_k)$ can be determined by proxy metrics. However, their values are hard to determine in practice, in particular, considering balancing the scales between backward and forward contribution scores. In our experimental results (Section~\ref{sec:Two Stage}), we find that manually scaling the scores results in suboptimal solutions.

To address this issue, we introduce a heuristic solution by two-stage decoupling. First, we decouple the multi-knapsack problem into sub-problems, each aiming to solve the orchestration problem on one device:
%\by{rewrite $\mathds{1}_{\mathscr{p}p_f}^{x_i}$  to $\mathds{1}_{\mathscr{p}p_f}(x_i)$}
\begin{align}
\label{eq:sub_problem}
\max&\sum_{i}^{N}\mathds{1}_{\mathscr{p}p_f}(x_i) \mathcal{A}^{p_f}(\mathcal{F}_k) + \mathds{1}_{\mathscr{p}p_o}(x_i) \mathcal{A}^{p_o}(\mathcal{F}_k), (\forall k)\\
s.t. & \sum_{i}^{N} \mathds{1}_{\mathscr{p}p_f}(x_i)(c^f_k + c^b_k) + \mathds{1}_{\mathscr{p}p_o}(x_i) c^f_k \leq \bm{C}_k,\nonumber\\
& \mathds{1}_{\mathscr{p}p_f}(x_i) + \mathds{1}_{\mathscr{p}p_o}(x_i) \leq 1, \quad \mathds{1}_{\mathscr{p}p_f}(x_i), \mathds{1}_{\mathscr{p}p_o}(x_i) \in \{0,1\}\nonumber.
\end{align}

Second, we decouple each sub-problem into a bi-level optimization problem where the outer problem optimizes the contribution regarding backward contribution scores under the Full operation computational cost $C^{p_f}_k$, and the inner problem optimizes the contribution regarding forward contribution scores. Therefore, the problem becomes:
\begin{align}
\label{eq:backward opt}
\max_{\mathds{1}_{\mathscr{p}p_f}(x_i)} & \sum_{i}^{N}\mathds{1}_{\mathscr{p}p_f}(x_i) \mathcal{A}^{p_f}(\mathcal{F}_k) \quad \text{(outer-level)}\\
s.t. & \sum_{i}^{N}\mathds{1}_{\mathscr{p}p_f}(x_i)(c^f_k + c^b_k) \leq \bm{C}_k, \nonumber\\
&\mathds{1}_{\mathscr{p}p_f}(x_i) + \mathds{1}_{\mathscr{p}p_o}(x_i) \leq 1, \nonumber\\
& \mathds{1}_{\mathscr{p}p_o}(x_i) \in \mathcal{S}, \quad \mathds{1}_{\mathscr{p}p_f}(x_i) \in \{0, 1\} \nonumber
\end{align}
with
\begin{align}
\label{eq:forward opt}
\mathcal{S} = \arg\max {\mathds{1}_{\mathscr{p}p_o}(x_i)} & \Bigl\{\sum_{i}^{N}\mathds{1}_{\mathscr{p}p_o}(x_i) \mathcal{A}^{p_o}(\mathcal{F}_k) \quad \text{(inner-level)} \\
s.t. & \sum_{i}^{N}\mathds{1}_{\mathscr{p}p_o}(x_i)c^f_k \leq \bm{C}_k, \nonumber\\
&\mathds{1}_{\mathscr{p}p_o}(x_i) \in \{0, 1\} \Bigl\}.\nonumber
\end{align}

By decoupling the orchestration problem in a hierarchical manner, both outer-level and inner-level subproblems can be solved independently, via the classic knapsacking algorithm, e.g., dynamic programming~\cite{pferschy1999dynamic,bertsimas2002approximate}. 

\subsection{Proposed heuristic algorithm for D2FT}
We detail the heuristic algorithm in Algorithm~\ref{Alg: Overall} and~\ref{Alg: DP}. Algorithm~\ref{Alg: Overall} shows the basic flow of the algorithm, and the algorithm~\ref{Alg: DP} presents the dynamic programming solution on the bi-level optimization problem we proposed in section~\ref{sec:Heuristic Algorithm}. The output of this DP algorithm is the scheduling table containing the operation selection $\mathds{1}_{\mathscr{p}p_f}(x_i)$ and $\mathds{1}_{\mathscr{p}p_o}(x_i)$ for $p_f$ and $p_o$ of all samples. When training, subnets will follow this scheduling table to process every sample. We denote scheduling tables $T_{opt}^{p_f}$ for $p_f$ and $T_{opt}^{p_o}$ for $p_o$. After obtaining $T_{opt}^{p_f}$ and $T_{opt}^{p_o}$, we merge the results into $T_{opt}$. Samples selected by $T_{opt}^{p_f}$ perform $p_f$ with a corresponding value of $1$ in $T_{opt}$ ($T_{opt}[k][x_i] = 1$). Samples from $T_{opt}^{p_o}$ are assigned $p_o$ with a value of 2 ($T_{opt}[k][x_i] = 2$). If a sample has been chosen in both $T_{opt}^{p_f}$ and $T_{opt}^{p_o}$, it performs $p_f$, and we set its value in $T_{opt}$ as 1. Samples not included in either in $T_{opt}^{p_f}$ or in $T_{opt}^{p_o}$ are scheduled to perform $p_s$, with a value of $3$ in $T_{opt}$ $T_{opt}[k][x_i] = 3$.
%\begin{minipage}{.49\columnwidth}
\begin{algorithm}
\SetAlgoLined
\SetKwInOut{Input}{Input}\SetKwInOut{Output}{Output}
% \SetKwInOut{Initialization}{Initalize}
\caption{KnapsackScheduling}
\label{Alg: Overall}
\Input{The forward contribution scores $\{\mathcal{A}^{p_o}\}$ and backward contribution scores $\{\mathcal{A}^{p_f}\}$ for all subnets $\mathcal{F}_k$ with each score list as $\mathcal{A}^{p_o}(\mathcal{F}_k)$ and $\mathcal{A}^{p_f}(\mathcal{F}_k)$;
The forward computational costs $c^f_k$ and backward computational costs $c^b_k$ for subnet $k$; 
The total device number $K$, computational capacity $C_k$ for each device, including $C^{p_f}_k$ for Full operation and $C^{p_o}_k$ for Forward-Only operation;
Dataset $\mathcal{D} = \{x_i\}$, with total $N$ data samples.}
\Output{A scheduling table $T_{opt}$.} 
%\by{you haven't described what a scheduling table is? How does it look like?} for all subnets to achieve workload balancing. }
Initialize Full operation ($p_f$) costs $W_b$, and Forward-Only ($p_o$) costs $W_f$;\\
%columns indicating a subnet, rows indicating a subnet.\\ 

%\% Assign cost for each subnet $\mathcal{F}_k$ on each sample $x_i$ in $D$. \\
\%Assign cost for each subnet $\mathcal{F}_k$. Columns in $W_b$ and $W_f$ indicate a subnet, and rows indicate a subnet.\\
\For{subnet $\mathcal{F}_k$, $k \in K$}{ 
   \For{$ i = 1, \ldots, N$}{ 
       $W_f[k][i] = c^f_k$ \\
       $W_b[k][i] = c^f_k + c^b_k$\\
   }
}

\% \textbf{Schedule $p_f$ and $p_o$ using Algorithm~\ref{Alg: DP}}.\\
$T_{opt}^{p_f}$ = DPSearching($\{\mathcal{A}^{p_f}\}$, $W_b$, $C^{p_f}_k$, $c^f_k$, $c^b_k$)\\ %\by{why two parentheses?}\\
$T_{opt}^{p_o}$ = DPSearching($\{\mathcal{A}^{p_o}\}$, $W_f$, $C^{p_o}_k$, $c^f_k$, $c^b_k$)\\
%\by{write the algorithm below using If operation, do you have else condition?}
Initialize scheduling table $T_{opt}$ as a zero matrix, with rows for samples and columns for subnets.\\
Each element in $T_{opt}$ can be $1$ (perform $p_f$), $2$ (perform $p_o$), or $3$ (perform $p_s$).\\
% The item values of $T_{opt}$ are represented as follows: $1$ denotes performing $p_f$, $2$ denotes performing $p_o$, and $3$ denotes performing $p_s$.\\
\For{subnet $\mathcal{F}_k$, $k \in K$}{
   \For{$ i = 1, \ldots, N$}{
\If{ only $T_{opt}^{p_f}[k][x_i] = 1$}{
    $T_{opt}[k][x_i] = 1$
}
\If{ only $T_{opt}^{p_o}[k][x_i] = 1$}{
    $T_{opt}[k][x_i] = 2$
}
\% Perform $p_f$ when device $k$ selects sample $x_i$ for both $p_f$ and $p_o$.\\
\If{$T_{opt}^{p_f}[k][x_i] = 1$ and $T_{opt}^{p_o}[k][x_i] = 1$}{
    $T_{opt}[k][x_i] = 1$
}
\% Perform $p_s$ when device $k$ chooses sample $x_i$ for neither $p_f$ nor $p_o$.\\
\If{$T_{opt}^{p_f}[k][x_i] = 0$ and $T_{opt}^{p_o}[k][x_i] = 0$}{
    $T_{opt}[k][x_i] = 3$
}
}
}

\textbf{Return} The scheduling table $T_{opt}$. 
\end{algorithm}
%\end{minipage}
%\begin{minipage}{.49\columnwidth}
\begin{algorithm}
\SetAlgoLined
\SetKwInOut{Input}{Input}\SetKwInOut{Output}{Output}
% \SetKwInOut{Initialization}{Initalize}
\caption{DPSearching}
\label{Alg: DP}
\Input{The set of contribution score list $\{\mathcal{A}^{p}\}$ of operation $p$ for all subnets. For subnet $k$, its contribution score list is $\mathcal{A}^{p}(\mathcal{F}_k)$; The computational cost list $W_{p}$ of operation $p$; The operation computational cost capacity $C^{p}_k$ for all subnets; The forward computational cost $c^f_k$ and backward computational cost $c^b_k$; Number of data samples $N$.}
\Output{A scheduling table $T_{opt}^{p}$ for operation $p$.}
Initialize a zero matrix searching table $T$, where $k$ indexes subnets and $i$ indexes samples in $\mathcal{A}^{p}(\mathcal{F}_k)$. Set $w$ as the capacity limitation for searching. \\
Initialize $T_{opt}^{p}$ as a zero matrix, with rows for samples and columns for subnets. $p$ indicates the selection of operations, including $p_f$,$p_o$.\\
Each element in the table $T_{opt}^{p}$ can be $0$ (not select), or $1$ (select).\\

\% Phase 1: Searching for maximum contribution scores.\\
\For{subnet $\mathcal{F}_k$, $k \in K$}{
    %$Score = \{\mathcal{A}^\mathscr{p}(\mathcal{F}_k)\}$\\
    \For{item $i = 1, 2 \ldots$ N}{
        \For{capacity $w = 0,1 \ldots C^{p}_k$}{
           \If{$w \geq (c^f_k + c^b_k)$}{
                $T[k][i][w]$ = max($T[k][i - 1][w]$,\\
                   \qquad $T[k][i - 1][w - c] + \mathcal{A}^{p}(\mathcal{F}_k)[i-1]$)\\
            }
            \Else{
                $T[k][i][w] = T[k][i - 1][w]$\\
            }
        } 
    }
}
\% $T[k][N][C^{p}_k]$ indicates the maximal contribution scores for index $k$ in Equation~\ref{eq:backward opt} and~\ref{eq:forward opt}.

% max value of $T$ for every index $k$ is the item $T[k][len(Score)][C_k]$.
% \% Maximal value of a subnet $\mathcal{F}_k$ is derived from $T$, denoted as $V_k$.

% The max value in $T$ for a subnet $\mathcal{F}_k$ is the maximize $V_k$. The max value of $T$ for every index $k$ is the item $T[k][len(Score)][C_k]$.\\ 
\vspace{0.5em}
\% Phase 2: Scheduling samples according to the maximum contribution scores.\\
%Subnet $\mathcal{F}_k$ max value $V_k = T[k][N][C_k]$\\
\For{subnet $\mathcal{F}_k$, $k \in K$}{
    $i = N, w = C^{p}_k$\\
    \While {$i > 0$ and $w > 0$}{
        \If {$T[k][i][w] \neq T[k][i - 1][w]$}{
            $T_{opt}^{p}[k][i-1]$ = 1\\
            $w = w - W_{p}[k][i - 1]$
        }
        $i = 1 - 1$
    }
}
\textbf{Return} Scheduling table $T_{opt}^{p}$ for operation $p$.
\end{algorithm}
\subsection{Extended D2FT for LoRA}
% \vspace{-0.3em}
\label{sec:lora}
In addition to conventional full parameter fine-tuning, the proposed D2FT framework can be generalized to parameter-efficient fine-tuning (PEFT) algorithms to enable computational-efficient fine-tuning further. In this work, we use low-rank adaptation (LoRA)~\cite{hu2021lora}, an intensively investigated PEFT algorithm, to demonstrate the generalization capabilities of D2FT. 
In the fine-tuning process, LoRA freezes most parameters in the neural network and only optimizes low-rank matrices in adapters, thereby significantly reducing computational costs.

To extend D2FT for LoRA-based fine-tuning, we follow the standard approach of applying LoRA updates to the query, key, and value (QKV) matrices in each attention head. Each LoRA matrix is co-located with its corresponding QKV matrix on the same device to maintain computational alignment. The overall partitioning strategy remains the same as described in Section~\ref{sec: partition strategy}, with the only difference being that each attention head is paired with its associated LoRA matrix. During fine-tuning, the foundation model parameters remain frozen, and D2FT is applied exclusively to the LoRA matrices. Experimental results (Section~\ref{sec: LoRA result}) confirm the effectiveness of D2FT in this LoRA-based setting.
\section{Experiments}
% \vspace{-0.2em}
\subsection{Experiment Settings}
\label{sec: setting}
\begin{figure}[!tb]
    \centering
    \includegraphics[width=0.99\linewidth]{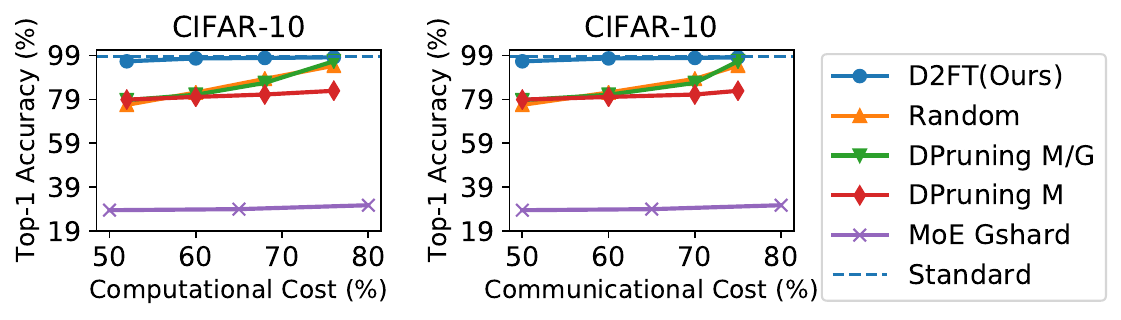}
    \caption{The top-1 accuracy comparison under the same or similar computation and communication costs on CIFAR-10 datasets.
    The ``Standard'' means the standard full fine-tuning.
    %From the lines shown in this figure, our method outperforms the baseline methods on Cifar-100 and Cifar-10 when the communication costs are close.
    }
    \label{fig: Experiment results Cifar-10}
    %\vspace{-1em}
\end{figure}

\noindent\textbf{Datasets.}
We chose CIFAR-10, CIFAR-100, and Stanford Cars as our test datasets. To improve efficiency, we use batch inputs and introduce micro-batches—subsets smaller than a full batch that serve as the evaluation unit. In our experiments, we replace individual samples with micro-batches. We set the batch size to $80$ for CIFAR and $25$ for Stanford Cars, with micro-batches of size $16$ and $5$, respectively, resulting in five micro-batches per batch. We also explore alternative micro-batch configurations in Appendix \ref{sec:Batch size}.

\noindent\textbf{Model.}
Our model is a ViT-small base model, pre-trained using timm~\cite{rw2019timm} for all experiments presented in this paper. The model consists of $12$ transformer blocks, each with $6$ attention heads and a feed-forward network comprising two fully connected layers. We use a patch size of $16$ and set the input dimensions to $224 \times 224$.

\noindent\textbf{Hardware and tools.} We use Tesla V100 GPU and use pytorch~\cite{paszke2019pytorch} tools to run each experiment. 

\noindent\textbf{Full fine-tuning partition settings.}
In our experiments, we split the ViT-small network into $74$ subnets: $72$ subnets correspond to the $12$ transformer blocks, with $6$ subnets representing each block. The remaining two subnets comprise the patch embedding layer at the start and the global pooling and classification layer at the network's end, respectively.
Each subnet corresponds to a block containing one attention head and $1/6$ of the feed-forward network. To exclude the impact of norm layers in forward and backward propagation, we freeze the parameter of norm layers in attention heads and replicate it for every subnet. Each subnet can deploy on a device.
%In real experiments, we simulate our settings by splitting the entire network into 74 subnets, 72 of which are related to the 12 ViT blocks. Each subnet belonging to a block applies a numerical gate to simulate device selections and operations. 
In the ablation study, we consider other splitting settings in appendix~\ref{sec:Device number}.
%other cases, $38$ and $26$ devices participating in the training. We enlarge the memory constraint of devices, assigning $36$ or $24$ devices to contain $12$ transformer blocks, each containing $3$ or $2$ devices. Therefore, each device contains $2$ or $3$ attention heads with $1/3$ and $1/2$ feed-forward networks, respectively. 
%We denoted the device containing one attention head and $1/6$ of the feed-forward network as ``$1/6$ block'', and the two attention heads and $1/3$ of the feed-forward network as ``$1/3$ block''. 

\noindent\textbf{LoRA fine-tuning partition settings.}
We split the ViT-small network into $74$ subnets. We assign $2$ subnets to represent the patch embedding layer at the beginning and the global pooling and classification layer at the end. The other $72$ subnets correspond to the $12$ transformer blocks and the LoRA matrices. Each subnet of blocks contains a frozen attention head, $6$ LoRA matrices for Query, Key, and Value matrices, and $1/6$ of the frozen feed-forward network. The gradients are only calculated for the LoRA matrices.

%The low-rank adaptation metrics and the feed-forward network are represented by $72$ subnets (We called computing subnet), each containing the ``AB'' metrics for the Query and Value components in LoRA and $1/6$ of the feed-forward network in which the parameter has been frozen. When the LoRA metric computes the results on the computing subnet, the freezing attention subnet receives the output from the computing subnets and sends the concatenated result to them. Then, the computing subnet sends the received results to its feed-forward network. Each subnet can be deployed on different devices. 

%. Six metric subnets represent the LoRA adaptation metrics for each block, with a rank of 96 for each metric. Additionally, 72 FFN subnets represent the feed-forward networks in ViT, each applying $1/6$ of the frozen feed-forward network, with six FFN subnets representing each block. The remaining two subnets consist of the patch embedding layer at the beginning and the global pooling and classification layer at the end of the network.
\noindent\textbf{Baseline methods for full fine-tuning.}
We consider four baseline methods applicable to distributed learning or sparsity fine-tuning, including Gshard MoE~\cite{lepikhin2020gshard}, two dynamic pruning methods~\cite{lin2020dynamic, sokar2022pay}, and random scheduling method. 
The dynamic pruning baseline methods change the pruning selection with $16$ iterations of fine-tuning (pruning according to the magnitude of each subnet~\cite{lin2020dynamic} or gradient of each subnet~\cite{sokar2022pay}), and the pruning method is the same as the subnet selection. We denote the magnitude dynamic pruning as ``DPruning M'' and the magnitude-gradient dynamic pruning as ``DPruning M/G''. The random scheduling, denoted as ``Random,'' randomly selects micro-batches for each subnet to do the $p_f$, $p_o$, and $p_s$ options. The Gshard method, denoted as ``MoE Gshard'', applies a gating network to schedule micro-batches to perform forward and backward propagation on different experts (experts deployed partial feed-forward network).  
The dynamic pruning and random scheduling share the same settings as D2FT. The Gshard methods apply $12$ subnets to comprise multi-attention heads for the $12$ blocks, with $3$ of attention heads for each. There are $5$ experts for each block ($60$ experts for $12$ blocks), and each expert contains $1/5$ of the feed-forward network. 
To achieve a better comparison, we also add the results of our experiment model standard fine-tuning (denoted as ``Standard'', which means all parameters participate in fine-tuning for all samples) as a baseline.

\noindent\textbf{Baseline methods for LoRA fine-tuning.}
We establish two LoRA fine-tuning baselines for comparison. Since the rank $R$ affects computational cost, we define a standard LoRA fine-tuning baseline with rank $R=240$, labeled as the ``Standard LoRA''. Additionally, we set ``LoRA w/ small rank'' baselines using standard LoRA fine-tuning with a smaller rank $R = 1$, $R = 60$, and $R = 200$. In configurations where only $3$ micro batches perform $p_f$ or $p_o$, two perform $p_s$, resulting in a total computation cost comparable to LoRA fine-tuning with $R = 1$. When $4$ micro-batches perform $p_f$ or $p_o$, one perform $p_s$, matching the computational cost of $R=60$.  When $5$ micro-batches perform $p_f$ and $p_o$, matching the computational cost of $R=200$. We will give the setting details in section~\ref{sec: LoRA result}. The ``Standard LoRA'' baseline emphasizes D2FT's computational efficiency with similar structures, while the ``LoRA w/ small rank'' baselines demonstrate how D2FT improves model performance at comparable computational costs.

\noindent\textbf{Evaluation metrics.} 
We consider five evaluation metrics to measure the performance of our proposed method.
\begin{itemize}[leftmargin=1em]

    \item \textbf{Top-1 accuracy:} We apply the top-1 accuracy to measure the model performance after fine-tuning.

    \item \textbf{Computational costs:} The computational costs measure the percentage of computation costs after applying D2FT or baseline methods compared with full fine-tuning or LoRA fine-tuning. 

    \item \textbf{Communicational costs:} The communicational cost evaluates the number of connections between devices after applying D2FT or baseline methods compared with full fine-tuning. %We present the results in appendix~\ref{sec: communication}

    \item \textbf{Workload variance:} The variance of workloads across devices in terms of computational or communicational cost.
    % The load of a subnet means the proportion of inputs sent to a subnet. 
    The higher variance indicates more imbalanced workloads.

    \item \textbf{Execution time:} Time for a subnet processing a batch input. We test the execution time to demonstrate the efficiency of fine-tuning methods.

\end{itemize}
We only consider the computation cost, communication cost, execution time, and load variance in fine-tuning process. In the inference, the model fine-tuned by D2FT uses all its parameters to predict the results. Therefore, we do not compare these in the inference phase. 
\vspace{-0.2em}
\subsection{Experiment Results}
\label{sec:results}
%\vspace{-0.5em}
\begin{table}[!tb]    
\centering
    % \small
    \caption{Comparison of workload variance. We present the variance in computation allocated to all devices. Lower variance indicates a better workload balance. The computational budget is set around $60\%$. The results show the optimal workload balance achieved by D2FT.}
    \vspace{-0.5em}
    %\resizebox{\linewidth}{!}{%
    \begin{tabular}{lr}
        \toprule
        \centering
        Methods & Workload Variance\\
        \midrule
        D2FT (Ours) & $0.00$\\
        Random & $0.23$\\
        DPruning M/G & $0.25$\\
        DPruning M & $0.25$\\
        MoE Gshard & $0.22$\\
        \bottomrule
    \end{tabular}
    %}
    \label{tab: Experiment results}
    % \vspace{-1.0em}
\end{table}
\begin{table}[!tb]
\centering
    % \small
    \caption{Comparison of execution time. We present the execution time for a single subnet processing assigned samples. The computational budget is set around $60\%$.}
    \vspace{-0.5em}
    % \resizebox{0.91\linewidth}{!}{%
    \begin{tabular}{lrr}
        \toprule
        \centering
        Methods & Execution Time & Top-1 Accuracy\\
        \midrule
        D2FT (Ours) & 28.80ms & $89.4\%$\\
        Random & 30.64ms & $44.4\%$\\
        DPruning M/G & 37.92ms & $52.6\%$\\
        DPruning M& 37.92ms & $54.4\%$\\
        MoE Gshard & 25.24ms & $39.1\%$\\
        \bottomrule
    \end{tabular}
    % }
    \label{tab: Experiment results1}
    % \vspace{-1em}
\end{table}
\begin{figure}[!tb]
    \centering
    \includegraphics[width=0.99\linewidth]{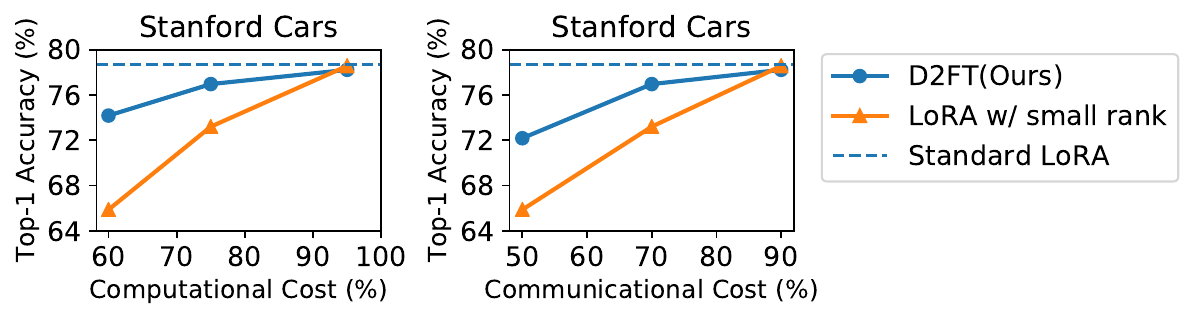}
    \vspace{-0.2em}
    \caption{Comparison of LoRA fine-tuning. We present the top-1 accuracy under the same or similar computational and communications costs on the Stanford Cars dataset. The ``Standard LoRA'' shows standard LoRA fine-tuning performance with the same rank as D2FT. The ``LoRA w/ small rank'' shows standard LoRA fine-tuning performance with a smaller rank to match D2FT's computational costs.  
    %From the lines shown in this figure, our method outperforms the baseline methods on CIFAR-100 and CIFAR-10 when the communication costs are close.
    }
    \label{fig: LoRA Experiment results}
    %\vspace{-1em}
\end{figure}
%\subsubsection{Model performance comparison}
In this section, we present the performance of models fine-tuned using D2FT compared to baseline methods, with top-1 accuracy comparisons conducted under similar computational and communicational costs.
Figures~\ref{fig: Experiment results Car} and~\ref{fig: Experiment results Cifar-10} illustrate these comparisons across three datasets, maintaining both consistent computation and communication costs across all methods. Results show that D2FT outperforms baseline methods under the same computational budget. Although our method experiences a slight accuracy drop compared to standard fine-tuning (e.g., $0.9\%$ on CIFAR-10, $2\%$ on CIFAR-100, and $2\%$ on Stanford Cars), it still demonstrates a $10\%$ improvement over other baseline methods on the CIFAR-10, CIFAR-100 datasets, and significantly outperforms all baseline methods on the Stanford Cars dataset.
%In summary, D2FT proves to be more effective for fine-tuning large, distributed pre-trained models compared to baseline methods, especially in scenarios with limited training resources, such as restricted computational or communication budgets. This efficiency makes D2FT particularly advantageous in resource-constrained settings, achieving high performance while minimizing resource demands.
%Figure~\ref{fig:subfig2} shows the comparison under the same communication costs. We choose points when all methods have the same or similar communication costs. We can conclude the same conclusion as Figure~\ref{fig:subfig1}, in which D2FT is worse than the normal fine-tuning but higher than all baseline methods on the two datasets.

\noindent\textbf{Comparison with Random scheduling.} 
D2FT achieves approximately $20\%$ higher top-1 accuracy on the CIFAR dataset and nearly $75\%$ higher on the Stanford Cars dataset compared to random scheduling in both computational and communication cost analyses. Since the primary distinction between our method and random scheduling is the scheduling algorithm, these results underscore the effectiveness and critical role of our scheduling strategy in optimizing model performance. \\
\noindent\textbf{Comparison with dynamic pruning.} The dynamic pruning method results are similar to random scheduling, but it is still ineffective in fine-tuning the model. 
Although these methods can dynamically change the selected subnets during iterations, they cannot overcome the problem when the pruning ratio is higher (no $p_o$ option like D2FT); the model performance is poor. \\
%When the computation costs or communication costs are less than $60\%$, the top-1 accuracy of models fine-tuned by dynamic pruning methods dramatically decreases. \\
%To mitigate the impact of a high pruning ratio but effectively decrease the computation and communication costs, we give the $p_o$ operation for subnets, whereas the dynamic pruning method only provides $p_f$ and $p_s$. The $p_o$ operation improves the model performance and could also decrease the computation costs and communication costs during fine-tuning.
\noindent\textbf{Comparison with Mixture of Experts (MoE).} In our setup, the MoE (Mixture of Experts) method underperforms, with fine-tuned model accuracy staying below $43\%$ on all datasets. This is not due to MoE's inherent limitations, as it excels in scaling large models rather than supporting distributed deployment. However, in our distributed setting, MoE's performance is inadequate. 

In summary, D2FT proves to be more effective for fine-tuning large, distributed pre-trained models compared to baseline methods, especially in scenarios with limited training resources, such as restricted computational or communication budgets. This efficiency makes D2FT particularly advantageous in resource-constrained settings, achieving high performance while minimizing resource demands. 
%D2FT greatly improves model performance compared to random scheduling. We increase the model performance by applying the $p_o$ operation compared to dynamic pruning methods. The MoE method is not suitable for distributed setting.

\subsubsection{Workload balancing comparison}
Our problem focuses not only on the fine-tuning of model performance but also on the workload balancing problem for every device. We choose the cases when the computation cost is equal to $60\%$ in the workload experiments, in which $3$ micro batches perform $p_f$ in each batch. We apply two metrics—workload variance and execution time—to assess the balance of task distribution. The workload variance results present are shown in Table~\ref{tab: Experiment results}, and the execution time results are shown in Table~\ref{tab: Experiment results1}. 

\noindent\textbf{Comparison of workload variance.} The workload variance measures the difference of samples assigned to subnets in a batch. A larger variance means a more unbalanced sample assignment. From the results in Table~\ref{tab: Experiment results}, we prove that our proposed method could balance assigned samples where the variance is $0$, which means every subnet receives the same number of samples to process. The other methods' variances, however, are not less than $0.2$, which means they cannot control the sample number assigned to every subnet.

\noindent\textbf{Comparison of execution time.} The execution time measures the running time for the entire network processing a batch input. From the execution time of different micro-batches processed by a single subnet, we conclude that a shorter execution time is relevant to a more balanced workload scheduling. The results in Table~\ref{tab: Experiment results1} show that D2FT needs $28.80$ms to finish processing a batch of input, which is faster than random scheduling and dynamic pruning. This result proves that D2FT scheduled a more balanced workload for every subnet than random scheduling and dynamic pruning. On the contrary, the execution time of D2FT is longer than that of the MoE method, which shows a different conclusion. 
This occurs due to the balancing mechanism in the Gshard approach, where experts skip micro-batches once they hit their processing limit to ensure equitable workload distribution among most experts. 
However, due to workload limitations imposed on subnets by this mechanism, certain samples that require processing are skipped. Therefore, Gshard's lower execution time comes from fewer samples processed, which also causes poor model performance.
%the poor model performance, their mechanism cannot assign samples to all experts (similar to the pruning methods), and its balancing mechanism causes some samples not to participate in fine-tuning, which makes the model ineffective.
In conclusion, D2FT achieves a more balanced workload than the baseline methods.

\subsubsection{Experiment results on LoRA}
\label{sec: LoRA result}
We use Stanford Cars as testing datasets, following our LoRA configurations. We evaluate three settings for computational costs: 
\begin{itemize}
    \item $3$ micro-batches perform $p_f$, and $2$ perform $p_o$, reducing the computational cost to $95\%$ of standard LoRA fine-tuning.
    \item $3$ micro-batches perform $p_f$, $1$ performs $p_o$, and $1$ performs $p_s$, reducing the computational cost to $75\%$.
    \item $3$ micro-batches perform $p_f$ and $2$ perform $p_s$, further reducing the computational cost to $60\%$.
\end{itemize}
We evaluate three settings for communicational costs:
\begin{itemize}
    \item $3$ micro-batches perform $p_f$, and $2$ perform $p_o$, with the communicational cost at around $90\%$ of standard LoRA fine-tuning.
    \item $3$ micro-batches perform $p_f$, $1$ performs $p_o$, and $1$ performs $p_s$, reducing the communicational cost to $70\%$.
    \item $2$ micro-batches perform $p_f$, $1$ perform $p_o$, and $2$ performs $p_s$, reducing the communicational cost to $50\%$.
\end{itemize}
The results, shown in Figure~\ref{fig: LoRA Experiment results}, indicate that D2FT adapts well to LoRA settings, achieving computational cost and communications cost savings while maintaining performance. 
%In the CIFAR-100 dataset experiment, D2FT fine-tuning improves model performance by about $2\%$ in Setting 3. In Settings 1 and 2, it achieves comparable performance to the LoRA w/ small rank, with a small gap compared to the standard LoRA. The experiment on CIFAR-100 shows that D2FT can perform better when computation resources are limited. We think the LoRA w/ small rank same as D2FT performance is because of the testing model pre-trained on ImageNet, which CIFAR-100 is a subset of. 
For the experiments of computational costs comparison, D2FT LoRA fine-tuning achieves close accuracy to the ``Standard LoRA'', with a drop of $4.2\%$ top-1 accuracy while using just $60\%$ of the computational cost for fine-tuning and drop and only a drop of $1.8\%$ while using $75\%$ computational costs. Meanwhile, it outperforms the ``LoRA w/ small rank'' baseline, boosting accuracy by $8.5\%$ at approximately $60\%$ of the total computational cost and by $3.6\%$ at around $75\%$. When the computational cost of both the ``LoRA w/ small rank'' baseline and our method exceeds $90\%$, the model performances of the ``Standard LoRA'', ``LoRA w/ small rank'', and D2FT align closely, appearing nearly identical.
We could find a similar conclusion for the experiments of communicational cost comparison. D2FT demonstrates outstanding performance while significantly reducing communication costs. When utilizing only $50\%$ of the communication resources, D2FT achieves a top-1 accuracy that is merely $6.3\%$ below the ``Standard LoRA''. As the communication cost increases to $70\%$, this accuracy gap narrows to just $4.6\%$. More importantly, D2FT substantially outperforms the ``LoRA w/ small rank'', enhancing top-1 accuracy by $5.7\%$ at $50\%$ communicational cost and $3.7\%$ at $70\%$. When communicational cost increases by around $90\%$, the performance of ``Standard LoRA'', ``LoRA w/ small rank'', and D2FT are close.
%These results highlight D2FT's ability to effectively balance computational efficiency with high model performance in distributed learning environments. 
In summary, our proposed D2FT enhances training efficiency with only minimal accuracy reduction in LoRA settings, especially when computation resources are limited.

\subsubsection{Different option of backward and forward scores}
\label{sec: score candidate}
%\begin{table}[!tb]    
%\centering
%    \small
    %\vspace{-0.5em}
%    \caption{D2FT LoRA fine-tuning results with different kinds of backward/forward scores. If Backward = $FI$, the backward score uses Fisher information, and the forward score uses weight magnitude. If Backward = $M$, the backward score uses weight magnitude, and the forward score uses Fisher information.}
    % $65$ devices, $56$ $1/6$ block and $9$ $1/3$ block devices}
%    \vspace{-0.5em}
    %\resizebox{\linewidth}{!}{
%    \begin{tabular}{c|cc}
%        \toprule
%        \centering
%        Score candidates & Top-1 accuracy & Dataset\\
%        \midrule
%        Backward = $FI$ & $85.7\%$ &\multirow{2}{*}{CIFAR-100}\\
%        Backward = $M$ & $89.4\%$ &\\
%        \midrule
%        Backward = $FI$ & $52.9\%$ &\multirow{2}{*}{Standford Car}\\
%        Backward = $M$ & $75.1\%$ &\\
%        \bottomrule
%    \end{tabular}
    %}
%    \label{tab: Score candidate Lora}
%\end{table}

\begin{table}[!tb]    
\centering
    \small
    %\vspace{-0.5em}
    \caption{Impact of measurements for backward and forward contribution scores. We select different measurements for contribution scores in D2FT full fine-tuning on the Stanford Cars dataset. We tested eight combinations of backward and forward scores to identify the scoring method that yields the best model performance.}
    % $65$ devices, $56$ $1/6$ block and $9$ $1/3$ block devices}
    \vspace{-0.5em}
    %\resizebox{\linewidth}{!}{
    \begin{tabular}{llr}
        \toprule
        \centering
        Backward score & Forward score & Top-1 accuracy\\
        %\midrule
        %Backward = $FI$ & $83.6\%$ &\multirow{2}{*}{CIFAR-100}\\
        %Backward = $M$ & $89.7\%$ &\\
        \midrule
        \textbf{Weight Magnitude} & \textbf{Fisher Information} & \textbf{79.8\%}\\
        Fisher Information & Weight Magnitude &$75.8\%$\\
        \midrule
        Weight Magnitude & Gradient Magnitude & $79.0\%$\\
        Gradient Magnitude & Weight Magnitude &$71.5\%$\\
        \midrule
        Fisher Information & Taylor Importance &$74.9\%$\\
        Taylor Importance & Fisher Information & $74.6\%$\\
        \midrule
        Weight Magnitude & Taylor Importance &$79.7\%$\\
        Taylor Importance & Weight Magnitude & $75.4\%$\\
        \bottomrule
    \end{tabular}
    %}
    \label{tab: Score candidate full}
    %\vspace{-0.5em}
\end{table}

We propose using Fisher Information, Weight Magnitude, Gradient Magnitude, and Taylor Expansion-based Weight Importance as scoring metrics in the main content. We empirically chose Weight Magnitude as the backward score and Fisher Information as the forward score for previous experiments. This section presents experimental results evaluating various backward and forward score options in full fine-tuning and explains our chosen selections for each. In this experiment, we suppose each input batch consists of $5$ micro-batches, $2$ micro-batches perform $p_f$, $1$ micro-batch perform $p_s$, and $2$ micro-batches perform $p_o$. The experiment results are in table~\ref{tab: Score candidate full}, which indicate that using Weight Magnitude as the backward score yields better performance in full fine-tuning. A likely reason is that the pre-trained weights play a crucial role in downstream tasks and benefit from updates, so prioritizing weights with larger magnitudes ensures the model's performance. After determining the backward score, we find that when Fisher Information becomes the forward score, the fine-tuned model performance achieves the best performance. Therefore, we choose Weight Magnitude as the backward score and Fisher Information as the forward score. 

% \vspace{-0.5em}
\section{Ablation Study}
\label{sec:ablation}
\subsection{Analysis of computational and communicational cost}
This subsection presents the computational and communicational cost of different operation selections.\\ 
\noindent\textbf{Computational cost analysis.} To estimate the overall network execution time, we measure the time it takes for a single subnet to perform $p_f$ and $p_o$. Specifically, we evaluate the execution time for a subnet processing between 1 and 5 micro-batches. The results of these tests are presented in Table~\ref{tab: unit execution time}. We use execution time measurements to estimate the computational costs for Full($p_f$) and Forward-Only($p_o$) operations. Based on the data in Table~\ref{tab: unit execution time}, we observe that the forward computational cost is consistently around $40\%$ of the total forward-backward computational cost, irrespective of the number of micro-batches. Consequently, we assume that the $p_o$ option consumes $40\%$ of the computational cost of $p_f$, while the $p_s$ option incurs no computational cost. For all experiments, we employ an SGD optimizer with momentum. Using these assumptions, we calculate the computational cost metric to evaluate the efficiency of our method.
%and this ratio remains consistent regardless of the number of micro-batches. Therefore, we assume the $p_o$ option only uses $40\%$ computational cost of the $p_f$, and we assume if we choose the $p_s$ option, the computational cost is $0$. Our optimizer is an SGD optimizer with momentum, which we use in all our experiments. We use our assumption to calculate the computational cost metric in our experiments.
%The execution time measurement results show as follows:
%\begin{table}[!tb]    
%\centering
%    \small
    %\resizebox{\linewidth}{!}{%
%    \caption{The execution time for forward and forward-backward propagation.}
%    \vspace{-0.5em}
%    \begin{tabular}{c|r}
%        \toprule
%        \centering
%        Computing Options & Average Execution Time\\
%        \midrule
%        Forward-Backward (p_f) & $48ms$\\
%        Forward Only (FO) & $20ms$ \\
%        \bottomrule
%    \end{tabular}
    %}
%    \label{tab: execution time}
%    \vspace{-1.5em}
%\end{table}  

\noindent\textbf{Communicational cost analysis.} Each block's output in a transformer base model has the same size because the structure and dimensions of each block are identical. Meanwhile, the communicational cost during backward propagation is identical to that during forward propagation because the tensor sizes are the same during forward and backward. Therefore, if a device chooses the $p_o$ for a micro-batch, it can save $50\%$ of its communicational, and there will be no communicational if it chooses the $p_s$ for a micro-batch. We use this to calculate the communicational cost metric in our experiments.\\

\begin{table}[!tb]
\centering
    \caption{The execution time for a subnet when processing a different number of micro-batch.}
    \vspace{-1.0em}
    %\resizebox{\linewidth}{!}{%
    \begin{tabular}{crr}
        \toprule
        \centering
        Micro-batch Number & Execution Time ($p_f$) & Execution Time ($p_o$)\\
        \midrule
        $1$ & $2.01ms$ & $0.86ms$\\
        $2$ & $2.20ms$ & $1.01ms$\\
        $3$ & $2.27ms$ & $1.05ms$\\
        $4$ & $2.74ms$ & $1.20ms$\\
        $5$ & $3.16ms$ & $1.48ms$\\
        \bottomrule
    \end{tabular}
    %}
    \label{tab: unit execution time}
    % \vspace{-0.5em}
\end{table}
%\by{analysis on other topics?}
\subsection{The impact of the number of subnets on D2FT}
\label{sec:Device number}
In this section, we investigate whether the number of subnets affects the fine-tuning performance of D2FT. We introduce two additional subnet configurations: $38$ subnets and $26$ subnets, in comparison to the 74 subnets used in the previous experiment. The batch size and micro-batch size are kept consistent with the prior setup, with each subnet selecting 2$p_f$ and 2$p_o$ for micro-batches. The CIFAR-100 dataset is used for testing, with a batch size of $80$ and micro-batch size of $16$. The results of this comparison are summarized in Table~\ref{tab: Number of devices}. 

The results indicate that reducing the number of subnets decreases the fine-tuned model's performance at the same computation ratio. This reduction occurs because fewer subnets mean each device's subnet contains more parameters. For instance, when selecting $p_o$ or $p_s$ in configurations with $38$ or $26$ subnets, the parameters affected are double and triple, respectively, compared to the $74$-subnet case. These findings demonstrate that under equal computational capacity across devices, D2FT performs better with a larger number of subnets, allowing for finer granularity in parameter updates.

\subsection{The impact of micro-batch size on D2FT}
\label{sec:Batch size}
In this section, we examine whether the size of micro-batches affects the fine-tuning performance of D2FT. We evaluate three micro-batch sizes: $4$, $8$, and $16$, to analyze their impact. Given a batch size of $80$, these micro-batch sizes correspond to $20$, $10$, and $5$ micro-batches per batch, respectively. The experiments are conducted on the CIFAR-100 dataset, with computation costs fixed at $64\%$. Specifically, $40\%$ of data samples perform $p_f$, $40\%$ data samples perform $p_o$, and $20\%$ data samples perform $p_s$. The results are shown in Table~\ref{tab: micro-batch}.
\begin{table}[!tb]
\begin{minipage}[b]{.49\textwidth}
\begin{center}
\centering
    % \small
    %\resizebox{\linewidth}{!}{%
    \caption{Impact of the number of subnets on D2FT.}%\by{on ..} .}
    \vspace{-0.5em}
    \begin{tabular}{cr}
        \toprule
        \centering
        Number of subnets & Top-1 accuracy\\
        \midrule
        74 & $89.7\%$\\
        38 & $89.4\%$\\
        26 & $88.8\%$\\
        \bottomrule
    \end{tabular}
    %}
    \label{tab: Number of devices}
    \vspace{0.5em}
\end{center}
\end{minipage}\hfill
\begin{minipage}[b]{.49\textwidth}
\begin{center}  
\centering
    % \small
    %\resizebox{\linewidth}{!}{%
    %\vspace{0.5em}
    \caption{Impact of micro-batch sizes on D2FT.}
    \vspace{-0.5em}
    \begin{tabular}{cr}
        \toprule
        \centering
        Micro-batch size & Top-1 accuracy\\
        \midrule
        $4$ & $90.1\%$ \\
        $8$ & $89.8\%$ \\
        $16$ & $89.7\%$ \\
        \bottomrule
    \end{tabular}
    %}
    \label{tab: micro-batch}
    % \vspace{-1.0em}
\end{center} 
\end{minipage}
\end{table}

The results indicate that top-1 accuracy remains consistent across different settings, suggesting that micro-batch size has a limited impact on performance. However, minor effects persist: as the micro-batch size increases, the fine-tuned model performance decreases slightly. This is likely because larger micro-batches may include less relevant samples within a subnet, whereas smaller micro-batches allow the algorithm to select more critical samples for each subnet, leading to better optimization. 

In summary, the micro-batch size has minimal impact on our method. While it may slightly influence the model's performance, any improvements or reductions are marginal and do not significantly affect overall outcomes.

\subsection{D2FT Performance under heterogeneous capabilities.}
\label{sec:Heterogeneous}
\begin{table}[!tb]
\begin{minipage}[b]{.49\textwidth}
\begin{center}   
\centering
    \small
    %\vspace{-0.5em}
    \caption{D2FT results on memory heterogeneity devices}
    % $65$ devices, $56$ $1/6$ block and $9$ $1/3$ block devices}
    \vspace{-0.5em}
    %\resizebox{\linewidth}{!}{
    \begin{tabular}{c|r}
        \toprule
        \centering
        Number of large memory devices & Top-1 accuracy\\
        \midrule
        $9$ & $89.7\%$\\
        $14$ & $89.8\%$ \\
        $19$ & $89.8\%$\\
        \bottomrule
    \end{tabular}
    %}
    \label{tab: Memory Heterogeneity}
    \vspace{0.5em}
\end{center}
\end{minipage}\hfill
\begin{minipage}[b]{.49\textwidth}
\begin{center}    
\centering
    \small
    \caption{D2FT results on computation heterogeneity devices}
    \vspace{-0.5em}
    %\resizebox{\linewidth}{!}{
    \begin{tabular}{c|r}
        \toprule
        \centering
        Number of high speed devices &Top-1 accuracy\\
        \midrule
        $9$ & $89.7\%$\\
        $14$ & $89.9\%$\\
        $19$ & $89.8\%$\\
        \bottomrule
    \end{tabular}
    %}
    \label{tab: Computation Heterogeneity}
    %\vspace{-0.5em}
\end{center} 
\end{minipage}
\end{table}

The experimental results in Section~\ref{sec:Device number} prompt us to consider scenarios involving devices with heterogeneous capabilities. We continue to use a batch consisting of 5 micro-batches, with each micro-batch containing 16 data samples.
We consider two types of heterogeneous experiments: memory heterogeneity and computational heterogeneity.
Tables~\ref{tab: Memory Heterogeneity} and~\ref{tab: Computation Heterogeneity} present the results for memory and computation heterogeneity experiments, respectively. 

\noindent\textbf{Memory heterogeneity}: Regarding memory heterogeneity, devices were categorized into two groups based on memory size: small and large. Small devices were configured to hold one attention head and a $1/6$ feed-forward network from a transformer block, while large devices accommodated two attention heads and a $1/3$ feed-forward network from a transformer block. We tested three configurations of large-memory device counts: $9$, $14$, and $19$. In each batch, each device performed $2p_f$ operations and $2p_o$ operations.
As shown in Table~\ref{tab: Memory Heterogeneity}, we observe that D2FT performs effectively even in cases with heterogeneous memory. The model's fine-tuned performance across three different settings is consistent and remains comparable to the performance achieved with homogeneous memory devices under the same computation cost (where the homogeneous top-1 accuracy is $89.7\%$). This indicates that D2FT is robust to memory heterogeneity.

\noindent\textbf{Computational heterogeneity}: Regarding computational heterogeneity, we defined two categories of device computation capabilities: slow and high speed. We fixed the memory usage for all devices, with each device containing one attention head and a $1/6$ feed-forward network in a transformer block. The total number of devices (subnets) used in the experiment was $74$.
A slow-speed device processes $2p_f$ and $2p_o$ operations in each batch, while a high-speed device can handle $3p_f$ and $1p_o$ operations. Three configurations were tested for the number of high-speed devices: $9$, $14$, and $19$. 
As shown in Table~\ref{tab: Computation Heterogeneity}, D2FT also performs well when devices with varying computational capabilities. 

In conclusion, neither memory nor computation heterogeneity negatively affects the performance of the fine-tuned model, suggesting that D2FT is adaptable and effective under both memory and computational heterogeneity.

\subsection{The impact of Forward-Only ($p_o$)}
\label{sec:Forward}
We evaluate the effectiveness of the Forward-Only ($p_o$) operation. The $p_o$ requires around $40\%$ of computational costs and $50\%$ communicational costs to the Full operation ($p_f$),  yet it plays a critical role in improving the fine-tuned model's performance. To test this, we fix the number of $p_f$ operations per subnet and vary the number of $p_o$ operations in a batch. The experiment uses a batch size of $25$, divided into $5$ micro-batches (each containing $5$ samples), with the Stanford Cars dataset. 
Each subnet is configured to perform $p_f$ on only $1$ micro-batch. Consequently, we examine five $p_o$ configurations, ranging from no micro-batches performing $p_o$ (with $4$ micro-batches perform $p_s$, denoted as $0p_o$) to all $4$ micro-batches performing $p_o$ (denoted as $4p_o$). This setup allows us to assess the impact of varying $p_o$ frequency on model performance. 

Based on the results in Table~\ref{tab: Forward effect}, we observe that $p_o$ operations significantly enhance performance. This improvement becomes more pronounced as the number of micro-batches performing $p_o$ increases, correspondingly reducing the number of micro-batches performing $p_s$. In summary, when computational resources are constrained, adopting the Forward-Only($p_o$) approach offers an effective trade-off, allowing for notable performance gains with only a modest increase in computational cost.   
%In conclusion, the $p_o$ can improve the model performance, especially when the computation cost is lower.
\begin{table}[!tb]
\begin{minipage}[b]{.49\textwidth}
\begin{center}
\centering
    \small
    \caption{The Forward-Pass ($p_o$) effectiveness}
    \vspace{-0.5em}
    %\resizebox{\linewidth}{!}{
    \begin{tabular}{lrr}
        \toprule
        \centering
        Forward setting & Computational cost & Top-1 accuracy\\
        \midrule
        $0 p_o$ & $20.0\%$ & $77.2\%$\\
        $1 p_o$ & $28.0\%$ & $77.4\%$\\
        $2 p_o$ & $36.0\%$ & $77.7\%$\\
        $3 p_o$ & $44.0\%$ & $78.9\%$\\
        $4 p_o$ & $52.0\%$ & $79.2\%$\\
        \bottomrule
    \end{tabular}
    %}
    \label{tab: Forward effect}
    \vspace{0.5em}
\end{center}\end{minipage} \hfill
\begin{minipage}[b]{.49\textwidth}
\begin{center}
\centering
    \small
    \caption{Bi-level scheduling effectiveness}
    \vspace{-0.5em}
    %\resizebox{\linewidth}{!}{
    \begin{tabular}{llr}
        \toprule
        \centering
        Optimization problem & $\lambda$ & Top-1 accuracy\\
        \midrule
        Bi-level & N/A &$89.7\%$\\
        Scaler & Max &$89.8\%$\\
        Scaler & Min &$75.2\%$\\
        Scaler & 0.2 & $77.7\%$\\
        Scaler & 0.1 & $76.3\%$\\
        \bottomrule
    \end{tabular}
    %}
    \label{tab: Two Stage}
    \vspace{-1.0em}
\end{center}\end{minipage}
\end{table}

\subsection{The necessity for two-stage decoupling}
\label{sec:Two Stage}
In the methodology section, we discussed employing two-stage decoupling, where a bi-level optimization problem is used to optimize $p_f$ and $p_o$. This section demonstrates the effectiveness of two-stage scheduling. As a baseline method, we consider balancing the scales of backward and forward scores. To achieve this, we introduce a scaling factor $\lambda$ to adjust the forward scores to match the scale of the backward scores. We refer to this approach as ``Scaler'' in our experiments. We give four options: $\lambda$ to make all forward contribution scores smaller than backward contribution scores (denoted as ``Max'' scaler), $\lambda$ to make all backward contribution scores smaller than forward contribution scores (denoted as ``Min'' scaler), $\lambda = 0.1$, and $\lambda = 0.2$. We specify a particular computation cost ($2p_f$,$2p_o$, and $1p_s$ for five micro-batches) to test the ``Scaler'' performance. The comparison results are shown in Table~\ref{tab: Two Stage}.

The ``Max'' scaler ensures that all forward contribution scores are smaller than their corresponding backward scores, aligning with the bi-level optimization framework used in two-stage decoupling. The results in Table~\ref{tab: Two Stage} indicate that the ``Max'' scaler and bi-level optimization yield comparable model performance. Conversely, the ``Min'' scaler for the "Scaler" approach performs worse than the bi-level optimization of the two-stage decoupling framework. Additionally, compared to constant scalers, two-stage decoupling achieves higher model performance. This demonstrates that two-stage decoupling is more effective than simply balancing the backward and forward score scales. While the performance of the ``Scaler'' method depends on the choice of $\lambda$ selection, the results in Table~\ref{tab: Two Stage} suggest that finding an optimal $\lambda$ is challenging.
%we conclude that there might exist scalers that make One-stage scheduling better than Two-stage scheduling, but it is hard to find such an appropriate scaler.
%We record the selection options for batches of these two scheduling algorithms. The Normal scheduling with ``Same'' scaler assigns $1$ micro-batch to do $p_f$ and $4$ micro-batches to do $FO$, under the same computation costs. The Two Stage scheduling assigns $2$ micro-batches do $p_f$, $2$ micro-batches do $FO$ and $1$ micro-batch do $p_s$. The updated parameters (number of micro-batches to do $p_f$ option) for Two Stage is more than Normal scheduling with ``Same'' scaler. Therefore, the Two-stage scheduling is needed and important to guarantee the model performance after fine-tuning.

%\subsection{Different kind of backward and forward scores}
%\label{sec: score candidate}
%\input{sec/Score_cand}
\section{Related Work}
Sparsely activated models and dynamic pruning, well-suited for distributed learning, help reduce computational load on individual devices during training but rarely address device load balancing. This challenge motivates our work, which aims to optimize performance in distributed settings by minimizing computation costs while effectively balancing workloads across devices.
%\by{add one sentence to connect the two topics with our work. why mentioning SAM and Dynamic pruning?}
%\vspace{-0.3em}
\subsection{Sparsely Activated Models}
%\vspace{-0.3em}
Sparsely activated models (SAMs)~\cite{shazeer2017outrageously} are introduced to reduce the training or fine-tuning cost of large-scale models. A model uses all its parameters for training, which is densely activated in general training or fine-tuning. SAMs choose subset models from the entire large-scale model for different inputs to decrease the training costs. To accomplish this, SAMs introduce Sparsely Gated Mixture-of-Experts (MoE), where certain layers, called MoE layers, are replicated to form multiple parallel subnets, each referred to as an "expert" within the larger model. A selection mechanism is then applied to activate only a subset of experts for training on a given task. This approach enables the network to selectively engage the most relevant experts for different downstream tasks rather than relying on the full model.
The scheme for selecting experts can be routed by a gate network~\cite{shazeer2017outrageously}. The gate network selects the top-$K$ relative ``experts'' according to the gate network weight for each expert. However, the gating network encounters a workload imbalance among experts. While this issue can be mitigated using techniques like workload balancing loss~\cite{lepikhin2020gshard, fedus2022switch}, limiting expert capacity (i.e., the number of samples an expert can process)\cite{lepikhin2020gshard}, implementing a policy scheduler\cite{nie2023flexmoe}, or splitting the top-$K$ gates into $K$ top-$1$ gates~\cite{yang2021m6}, these routing methods do not enhance the parameter efficiency of SAMs~\cite{zuo2021taming}, and therefore may not improve model performance. Alternative approaches like replacing gating mechanisms with hash-based selection~\cite{roller2021hash} or using random selection alongside knowledge distillation updates~\cite{zuo2021taming} can aid in balancing workloads and enhancing performance. However, these methods focus solely on the workload distribution across devices with MoE layers, without considering the memory or computation constraints of the devices used for partitioning.
%\vspace{-0.4em}
\subsection{Dynamic Pruning}
%\vspace{-0.4em}
%Although many techniques have been proposed recently to tackle the burden of high computational costs, they often only alleviate the problem to a limited extent.
%For example, low (or mixed) precision~\cite{nakandala2020incremental} often reduce the gradient and parameter representation from single-precision (32-bit) to half-precision (16-bit) or even int-precision (8-bit) to save memory. However, further precision reduction results in a significant performance drop. Gradient-based approaches, such as gradient checkpointing (or rematerialization)~\cite{griewank2000algorithm} and gradient accumulation~\cite{zhang2023adam,huang2023measuring}, only reduce the memory cost in the parameter updates, and cannot solve more severe memory challenges when the forward passing is even infeasible on a single device. Therefore, the methods related to sparse activities or sparse training have become popular.
% Sparse training is proposed to solve the intensive memory for large-scale networks~\cite{srinivas2017training}. A common method for implementing sparse computation involves reducing redundancy within a model, often achieved by pruning redundant neurons or weights to conserve memory. 
Dynamic pruning removes the redundant model weights to reduce computing costs during fine-tuning and compress the model after fine-tuning.
%One popular pruning technique is dynamic pruning. 
Unlike other pruning methods, such as those applied before fine-tuning~\cite{lee2018snip, wang2020picking, kohama2023single}, after fine-tuning~\cite{han2015learning,guo2016dynamic,alvarez2017compression,carreira2018learning}, or iteratively during fine-tuning~\cite{frankle2018lottery, zhang2021efficient, girish2021lottery}, which cannot recover the pruned weights, dynamic pruning computes a candidate mask that allows later pruning results cover the former pruning., thereby correct pruning errors during the fine-tuning process and minimize the drop in model accuracy. 
Lin~\cite{lin2020dynamic} et al. and Sokar~\cite{sokar2022pay} et al. dynamically update network connections according to an importance score derived from batch gradients and weight magnitudes. Dynamic pruning demonstrates that it’s unnecessary to use all input data to train every network weight. However, while dynamic pruning selects a subnet for specific inputs, in distributed learning scenarios, workload imbalance can cause certain partitioned devices to remain idle, wasting computational resources. Meanwhile, the purpose of compressing the model is to decrease the performance of the fine-tuned model unavoidably.

Compared with existing works, our proposed D2FT approach addresses computation, memory overload, and workload balancing while preserving fine-tuning performance. 
% We deploy the large neural network using a scheduling algorithm that arranges the number of inputs and operation selection for each partitioned device to balance computational costs. 
%\by{use one sentence to show D2FT aims to address this}
\section{Conclusion}
% \vspace{-0.5em}
In this paper, we present a new framework, D2FT, which can dynamically allocate learning operations across distributed devices and improve workload balances across devices during fine-tuning. We formulate operation scheduling as a multiple knapsack optimization problem and optimize the fine-tuning schedule strategy using dynamic programming. We evaluate our proposed framework on CIFAR-10, CIFAR-100, and Stanford Cars datasets with four baseline methods in full fine-tuning settings. We also consider parameter-efficient fine-tuning, extending D2FT in LoRA settings with same rank standard LoRA and smaller rank standard LoRA fine-tuning as baseline methods. Our experimental results show that D2FT outperforms all baseline methods in full and LoRA fine-tuning settings, especially in low computational and communicational costs. The findings demonstrate that D2FT improves training efficiency with minimal accuracy loss.
{
    \small
    \bibliographystyle{IEEEtran}
    \bibliography{related, bib/deep, bib/distributed_learning}
}

%\clearpage
%\appendix
%\onecolumn
%\input{sec/Algorithm}
\end{document}